\documentclass{article}

\usepackage[english]{babel}

\usepackage[margin=2.5cm]{geometry}

\usepackage{amsmath}
\usepackage{graphicx}
\usepackage[colorlinks=true, allcolors=blue]{hyperref}
\usepackage{setspace}
\usepackage{natbib}
\usepackage{siunitx}
\usepackage{float}
\usepackage{caption}
\usepackage{subcaption}
\usepackage{enumitem}
\usepackage{threeparttable}
\usepackage[affil-it]{authblk}
\usepackage[utf8]{inputenc}

\title{Predicting Hurricane Evacuation Decisions with Interpretable Machine Learning Models}

\author[1]{Yuran Sun\thanks{\normalsize Corresponding author. Postal Address: 1949 Stadium Rd, Gainesville, FL 32611, USA. Email Address: yuransun@ufl.edu}}
\author[2]{Shih-Kai Huang}
\author[1]{Xilei Zhao}

\affil[1]{\normalsize Department of Civil and Coastal Engineering, University of Florida}
\affil[2]{\normalsize Department of Emergency Management and Public Administration, Jacksonville State University}

\date{}

\begin{document}

\begin{singlespace}
\maketitle
\end{singlespace}

\hrule

\section*{Abstract}
\large
The aggravating effects of climate change and the growing population in hurricane-prone areas escalate the challenges in large-scale hurricane evacuations. While hurricane preparedness and response strategies vastly rely on the accuracy and timeliness of the predicted households’ evacuation decisions, current studies featuring psychological-driven linear models leave some significant limitations in practice. Hence, the present study proposes a new methodology for predicting households’ evacuation decisions constructed by easily accessible demographic and resource-related predictors compared to current models with a high reliance on psychological factors. Meanwhile, an enhanced logistic regression (ELR) model that could automatically account for nonlinearities (i.e., univariate and bivariate threshold effects) by an interpretable machine learning approach is developed to secure the accuracy of the results. Specifically, low-depth decision trees are selected for nonlinearity detection to identify the critical thresholds, build a transparent model structure, and solidify the robustness. Then, an empirical dataset collected after Hurricanes Katrina and Rita is hired to examine the practicability of the new methodology. The results indicate that the enhanced logistic regression (ELR) model has the most convincing performance in explaining the variation of the households’ evacuation decision in model fit and prediction capability compared to previous linear models. It suggests that the proposed methodology could provide a new tool and framework for the emergency management authorities to improve the estimation of evacuation traffic demands in a timely and accurate manner.

\noindent\textit{Keywords}: Hurricane evacuation, decision-making modeling, interpretable machine learning, nonlinearity detection


\section{Introduction}
\large
The population in the U.S. Coastline regions reached 94.7 million people (or 29.1\% of the total U.S. population) in 2017, up from 47.4 million people six decades ago, an increase of 99.8 percent \citep{USBureau}. Meanwhile, over 50\% of households live within 50 miles of the shoreline \citep{oceanecon}. Furthermore, the major population growth is within the Gulf of Mexico regions and the adjacent Atlantic regions, which are the most vulnerable areas exposed to hurricane risk \citep{USBureau}. Therefore, establishing planning-based evacuation strategies becomes urgent and essential for most coastal counties \citep{lindell2019routledge}. Disaster researchers have been closely monitoring this issue and striving to improve knowledge and technology \citep{sorensen2000hazard}. As \citet{sorensen2000hazard} has noted, one of the primary challenges in the modern era is enhancing prediction and forecasting to enable timely and effective responses. 

Studies on evacuation decision predictions and derivative traffic demand estimations have been booming since \citet{baker1991hurricane} published his systematic review of hurricane evacuation case studies between 1965 and 1990. The majority of the follow-up causal studies focused on three related areas, including (\textbf{1}) exploring the reasons facilitating or inhibiting people’s evacuation \citep{dow1998crying}, (\textbf{2}) examining the causalities between the predictors and evacuation decisions \citep{horney2010factors, lindell2005household}, and (\textbf{3}) developing comprehensive models to address the cause-effect structure \citep{gladwin1997warning, article, huang2017multistage}. 
While these studies utilizing bivariate or linear analysis methods \citep{huang2017multistage, huang2016leaves, lindell2012protective, lindell2018communicating} have contributed to the development of evacuation demand models \citep{lindell2013evacuation, murray2013evacuation}, they have tended to overemphasize the role of individuals' psychological processes in mediating the effects of information-seeking/-receiving and other contextual circumstances \citep{lindell2012protective, lindell2018communicating}. As a result, these studies are limited in several respects.

First, the previous linear or bivariate-based analyses are primarily used to test causality and are not optimized for prediction purposes \citep{article}. Therefore, these models, like other social science models, tend to have lower model fit and accuracy when used to forecast evacuation demand, probably due to their inability to automatically detect and capture interactions and nonlinearities \citep{zhao2020modelling}.
Next, the direct application of the psychological models is limited by the availability of psychological data and the consistency of the measuring instruments \citep{baker1991hurricane}, as well as the risk of overlooking the crucial impacts of the social and household contexts, such as resource availability \citep{schorr2015bridging, metaxa2018social}, social vulnerabilities \citep{ metaxa2018social,meyer2018previous}, and social capital \citep{metaxa2018social, fraser2022fleeing}, in estimating evacuation demands. 

Possible measures can be adopted to address these limitations. With the recent advancements in Artificial Intelligence (AI) and machine learning, there is a significant opportunity to utilize these cutting-edge technologies to capture complex relationships (e.g., nonlinearities and interactions) with the data \citep{zhao2021using}, leading to more accurate predictions. To tackle the potential issues of the black-box nature of machine learning models (e.g., lacking interpretability and trust \citep{rudin2019stop}), a recent effort in the machine learning community is to adopt interpretable machine learning models for high-stakes decision-making \citep{rudin2019stop, zhao2021using}. A prominent example is \citet{DUMITRESCU20221178}, which developed the penalized logit tree regression model, which incorporates nonlinear low-depth decision trees to improve the logistic regression model. The approach described in \citet{DUMITRESCU20221178} first involves building one-layer decision trees for each predictor. These trees partition the data into two groups based on each predictor, with each group assigned to one of the leaf nodes in the tree. Binary variables are created to represent each leaf node. Subsequently, two-layer decision trees are constructed for every possible pair of predictors, partitioning the data into three leaf nodes. Similarly, binary variables are generated for each of the leaf nodes. The ultimate model in the approach is a penalized logistic regression model that includes binary variables associated with the leftmost leaf node of each one-layer decision tree, as well as binary variables associated with the two leftmost leaf nodes of each two-layer decision tree. It attains satisfactory prediction performance, exhibiting significant improvement over the logistic regression model with original predictors and comparable results to black-box machine learning models like Random Forest, while preserving a high level of model transparency. 
Due to the advantages of \citet{DUMITRESCU20221178}, we have decided to apply this method to model hurricane evacuation decisions, which may support life-and-death decision-making in emergencies. 
However, in the context of hurricane evacuation decision modeling, the method proposed in \citet{DUMITRESCU20221178} may not be applied directly: 
(\textbf{1}) The logistic regression model solely comprises an extensive set of binary variables, posing a challenge for emergency planners to interpret the variable coefficients; (\textbf{2}) the model does not capture the trends in how a particular predictor contributes to the response variable, i.e., the causal effects; (\textbf{3}) the statistical significance of the nonlinearities (binary variables derived from decision trees) was not tested in \citet{DUMITRESCU20221178}, which means that non-significant nonlinearities might be included in the model and result in overfitting. From the variable selection aspect, reducing dependence on psychological variables and emphasizing social and household contextual variables can be achieved by employing suitable machine learning models. These models help explore the underlying relationships between social and household contextual variables and evacuation decisions.

To overcome the limitations and bridge the gaps mentioned above, this study aims at: (\textbf{1}) developing an interpretable machine learning model that enhances the logistic regression (used by most prior studies to model evacuation decisions) with low-depth decision trees, to automatically capture nonlinear effects (i.e., univariate and bivariate threshold effects) \citep{molnar2020interpretable} and thus achieve better model fit and accuracy. Compared with the penalized logit tree regression model developed by \citet{DUMITRESCU20221178}, the key innovations of the proposed model include representing nonlinearities using cutting points of the piece-wise model (a type of logistic regression model that allows for changes in the slope of the regression line at one knot that called the cutting point) and the interaction terms instead of a large set of binary variables. This approach can reflect the nonlinear causal effects of specific variables through their corresponding coefficients. Additionally, only statistically significant nonlinearities are included in the proposed model to reduce dimensionality and mitigate the overfitting problem. (\textbf{2}) Empirically developing an evacuation decision prediction model based on the proposed framework with the considerations of social and household contextual variables and then examining the model fit and accuracy against traditional models. (\textbf{3}) Sharing the new behavioral insights on how social and contextual variables would affect evacuation decisions through the interpretation of the proposed model. 

The following sections include: A thorough literature review of key predictors associated with evacuation decisions and the corresponding modeling techniques, the development of an enhanced logistic regression (ELR) framework, the re-analysis of the Hurricanes Rita and Katrina evacuation data with ELR, the validation of ELR by comparing it to the baselines models (the model without nonlinear effects and the model with psychological variables), and the summary of key findings for future studies.

\section{Literature Review}
\subsection{Predictors for Household's Evacuation Decisions}
Previous comprehensive review studies \citep{baker1991hurricane, huang2016leaves} divided diverse predictors for households’ hurricane evacuation decisions into four categories---social and household contexts, geographic characteristics, information-related items, and psychological-related variables. In general, social and household contexts include gender, age, household size, race, income, education level, previous experience, home structure, as well as available resources and preparedness levels; geographic variables are about the proximity to hazard-prone areas; information-related variables measure information sources and warning dissemination; variables of psychological-related variables include individuals’ perceptions and expectations to the resource-related concerns, threat conditions and exposures, and possible consequences \citep{baker1991hurricane, gladwin1997warning}. Most hurricane evacuation studies \citep{lindell2005household, article, huang2017multistage, Lazo-2015-Factors} either followed this framework or partially focused on one or two categories when developing their statistical or mathematical models.

Social and household contexts received the most discussions in previous hurricane evacuation studies but usually yielded none or marginal significance \citep{lindell2005household, gladwin1997warning, article, huang2017multistage, huang2016leaves, Lazo-2015-Factors, Sarwar2018, Lindell2011, doi:10.1061/(ASCE)NH.1527-6996.0000244,doi:10.1080/02732170701534226, doi:10.1080/00139157.2015.1089145}. Among them, females and mobile homeowners were more likely to evacuate, whereas minorities and homeowners usually hesitated to leave \citep{baker1991hurricane, gladwin1997warning, article, huang2017multistage, huang2016leaves}. Resource requirements and availabilities were less studied. Only a few studies \citep{Lazo-2015-Factors, Sarwar2018, Lindell2011, doi:10.1061/(ASCE)NH.1527-6996.0000244} found a significant effect of vehicle ownership on households’ evacuation decisions. In fact, resource-related variables were more likely to be discussed in evacuation logistic studies \citep{Sarwar2018}. Proximity to risk sources, no matter how it was measured, was one of the strongest predictors \citep{lindell2005household, gladwin1997warning, article,huang2017multistage, huang2016leaves,Lazo-2015-Factors,Sarwar2018,Lindell2011,doi:10.1061/(ASCE)NH.1527-6996.0000244} as it mirrored evacuees’ immediate (and maybe intuitive) responses to the threat exposures \citep{huang2017multistage, huang2016leaves}. Similarly, information-related variables, including official warnings and social or environmental cues, were vital \citep{article,huang2017multistage, huang2016leaves,doi:10.1080/02732170701534226, doi:10.1080/00139157.2015.1089145} as they were direct stimulation \citep{huang2017multistage, huang2016leaves}. For the psychological-related variables, perceived threat exposures and expected consequences reportedly had the main effects in the evacuation decision models in most studies \citep{gladwin1997warning, article, huang2017multistage, huang2016leaves,Lazo-2015-Factors, collins2021hurricane} as the process of which mental cognitions determined the behavioral decisions were consistent with the Elaboration Likelihood Model of Persuasion \citep{petty1986elaboration}. 

\subsection{Hurricane Evacuation Decision Modeling}
Predictors mentioned above have long been entered into advanced data analyses to reach the results with comprehensive concerns. Of the diverse modeling techniques, descriptive and regression analyses are the most common methods. One prototypical study of hurricane evacuation decisions could be traced back to Peacock and Gladwin’s 1997 study of Hurricane Andrew \citep{gladwin1997warning}.

The study made comparisons of evacuation percentages in different residential zones and used logistics regressions to analyze the contribution to the evacuation decision of the predictors. Since then, the mixed logistic regression models (random parameter logistic regression model) have been frequently employed to explain the correlations and effects between predictors and evacuation decisions, including studies of Hurricanes Lili \citep{lindell2005household}, Rita and Katrina \citep{huang2017multistage}, Ike \citep{article}, Sandy \citep{doi:10.1061/(ASCE)NH.1527-6996.0000244}, and Ivan \citep{Hasan-etal-2011-Behaviroal, Sarwar2018}. Moreover, two studies by Huang and his colleagues \citep{article, huang2017multistage} and Morss et al. \citep{UnderstandingPublicHurricaneEvacuationDecisionsandResponsestoForecastandWarningMessages} developed multistage regression analyses to clarify the effect routes among the predictors. Other than that, ordered probit models were hired in Cahyanto et al.’s study of tourists’ hurricane evacuation decisions \citep{CAHYANTO2014253} and Lazo et al.’s hypothetical study \citep{Lazo-2015-Factors}. Recently, \citet{AHMED2020100180} stepped forward and modeled the influence of social networks on evacuation decisions with zero-inflated Poisson, Tobit, linear regressions, and the multinomial logistic regression model.

Despite the diverse modeling methods used in previous hurricane evacuation decision studies, most existing hurricane evacuation decision models assume the linear and additive contributions of variables. However, hurricane evacuation decision-making can be nonlinear, multilayered, and much more complex \citep{gladwin2007social}. While many recent studies have considered this complex nature by using the crowd modeling methods \citep{vreugdenhil2015using} or the tree-based framework \citep{zhao2020modelling,xu2023predicting} to detect the nonlinear contributions, such studies are still in the embryo stages as the findings drawn from those pioneering studies are in need of further verification.

\section{Methods}
In concluding the literature review above, previous studies have underlined that logistic regression models could effectively capture the relation between predictors and the response variable (i.e., the evacuation decision) \citep{gladwin1997warning, Hasan-etal-2011-Behaviroal, Sarwar2018, doi:10.1061/(ASCE)NH.1527-6996.0000244}. Causally, these models might over-rely on the psychological variables in predicting evacuation decisions \citep{huang2017multistage}. In contrast, the psychological variables are subjective and vulnerable to expense, timely gathering, and consistency issues in data collection. On the other hand, many objective and easily-accessible social and household contextual variables (e.g., demographic and resource-related variables) reportedly have minor or indirect effects on evacuation decisions.

\begin{figure}[H]
  \centering
\includegraphics[width=1.0\textwidth]{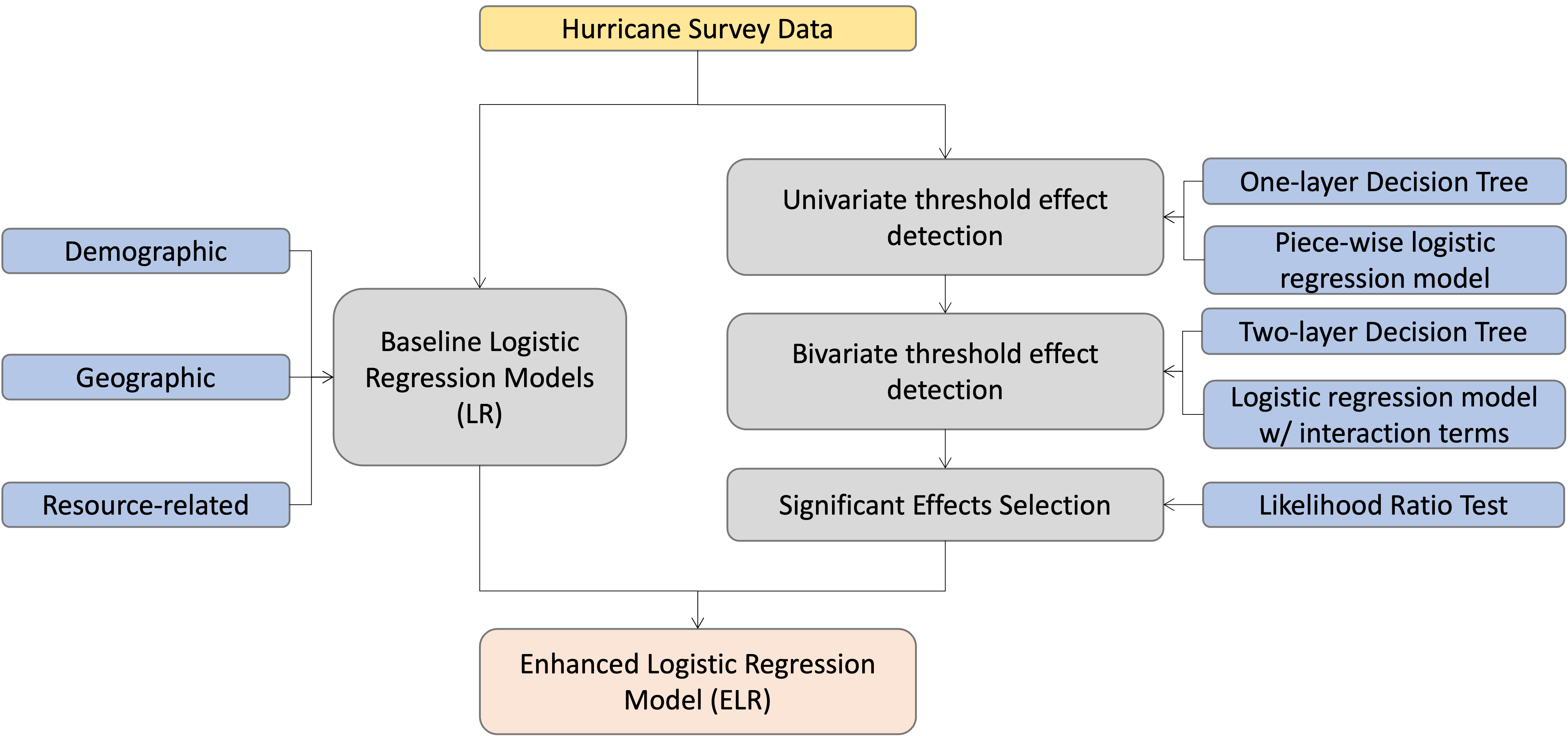}
  \caption{\small Overview of Methodology}
  \label{fig:flow}
\end{figure}

This study begins by proposing a framework indicating the establishment process of the interpretable machine learning model. To be specific, the framework assumes that the inconsistent contributions of the objective predictors can be attributable to the oversimple linear structure of logistic regression models. In other words, the study posits that such predictors might have crucial nonlinear effects on evacuation decisions. Thereby, as illustrated in Figure \ref{fig:flow}, this study develops a new methodology to (\textbf{1}) build a model with only the demographic, geographic, and resource-related predictors that could attain similar (or even better) predictive power as the logistic regression model with psychological variables; and (\textbf{2}) detect and model the nonlinearities with the interpretable machine learning approach \citep{Rudin2019} that is more flexible and intelligent than classic statistical methods (e.g., subjectively identifying nonlinear relationships from scatter plots) while retaining the model transparency.

In our study, we focus on two common nonlinearities——\textit{univariate threshold effects} and \textit{bivariate threshold effects}. The univariate threshold effect pertains to the varied impacts of a single predictor on a response variable as it moves across various ranges or levels. The bivariate threshold effect refers to the interaction effect of two predictors on a response variable, which exists only when the values of these two predictors fall within specific ranges or levels. We use a similar approach as \citet{DUMITRESCU20221178} to detect the threshold effects with low-depth decision trees with the following innovative modifications: (\textbf{1}) instead of using a vast number of binary variables in the logistic regression model like \citet{DUMITRESCU20221178}, incorporating the original predictors along with their nonlinearities, cutting points (which represent the thresholds for different intervals where a predictor may have varying impact on the response variable) and interaction terms identified by the one-layer and two-layer decision trees, in the logistic regression model, which makes it easy to identify and interpret the causal effects of specific variables by examining the resulting coefficients; (\textbf{2}) conducting statistical tests to determine the statistical significance of nonlinearities; and (\textbf{3}) retaining only the significant nonlinearities to reduce dimensionality and prevent the overfitting problem. Decision tree structures are ideal for capturing interactions between predictors, and they can visually display thresholds and classification results, thus offering good explanations \citep{molnar2020interpretable}. Particularly, low-depth decision trees are the most robust with straightforward explanations (since for each split, the observation falls into either one or the other leaf, and binary decisions are easy for humans to understand) \citep{molnar2020interpretable}; therefore, low-depth decision trees are suitable for threshold effects detection.

As shown in Figure \ref{fig:flow}, which outlines the overall framework of the study, we first build baseline logistic regression models (LR) framed among demographic, geographic, socioeconomic variables. Next, we detect \textit{univariate threshold effects} and \textit{bivariate threshold effects} using low-depth decision trees. We then select those significant threshold effects that passed the likelihood ratio tests and create an enhanced logistic regression model (ELR) incorporating all significant threshold effects.

In the following subsections, we explain the details of threshold effects, baseline, and enhanced logistic regression models, how low-depth decision trees can detect threshold effects, and statistical tests.

\subsection{Threshold Effects}
This paper focuses on two common nonlinear relationships among evacuation decisions and predictors: Univariate and bivariate threshold effects. The univariate threshold effect is the threshold effect of a single predictor. Denote the predictor as $x_i$, where $i \in \{1, 2, \cdots, p$\} and $p$ is the number of predictors. A univariate threshold effect exists if the relations between $x_i$ and the response variable (evacuation decision) are different when $x_i$ is below and above a certain constant threshold $a_i$. This can be seen through the differing coefficients of $x_i$ in the logistic regression model when $x_i \leq a_i$ and $x_i > a_i$ (the logistic function is shown in Section 3.2).

The bivariate threshold effect is the threshold effect of the interaction term of two predictors. Denote two predictors as $x_i$ and $x_j$, where $i, j \in \{ 1, 2, \cdots, p \}$. $x_i$ and $x_j$ have the bivariate threshold effect if their interaction term (i.e., $x_i \cdot x_j$) affects the evacuation decision when the values of $x_i$ and $x_j$ fall in one of the following intervals: $\textbf{(1)}$ $x_i \leq b_i$ and $x_j \leq b_j$; $\textbf{(2)}$ $x_i \leq b_i$ and $x_j > b_j$; $\textbf{(3)}$ $x_i > b_i$ and $x_j \leq b_j$; and $\textbf{(4)}$ $x_i > b_i$ and $x_j > b_j$. (the logistic function is shown in Section 3.2). 

Even though some within-category interactions could be anticipated, those interactions are usually minor. For example, households' socioeconomic status might be able to moderate the effect of household sizes on evacuation decisions. In contrast, resource availability and needs are better moderators than socioeconomic contexts in interpreting such effects. Hence, our study assumes the bivariate threshold effects occur between predictors from different categories rather than within categories, while within-level bivariate threshold effects are not considered in our model.

\subsection{Incorporating Threshold Effects into Logistic Regression Models}
Denote the evacuation decision as a binary variable $y_k\in\{0,1\}$ and the input data for $k_{th}$ observation as a p-dimensional vector \textbf{$x_k$}, where $p$ is the number of predictors, $k= 1, 2, \cdots, n$, and $n$ is the number of observations. The logistic regression model is built to estimate the probability $P(y_k=1|x_k)$ that a $k_{th}$ respondent decides to evacuate during the hurricane given individual's features $x_k$. An individual is deemed to evacuate when this probability is larger than a threshold $\pi$ and to stay at home otherwise. The logistic function of the baseline logistic regression model (LR) is:
\begin{equation}
P(y_k=1|x_k,\beta)=\frac{1}{1+e^{-(\beta_0+\sum_{i=1}^{p}\beta_ix_{k,i})}},
\end{equation}
where $\beta= (\beta_0,\beta_1,\cdots,\beta_p)^{T}$ is the set of parameters to be estimated. The optimum estimator $\hat{\beta}$ can be found by maximizing the log-likelihood function
\begin{equation}
L(\beta)=\sum_{k=1}^{n}y_kln(P(y_k=1|x_k,\beta))+\sum_{k=1}^{n}(1-y_k)ln(1-p(y_k=1|x_k,\beta)).
\end{equation}
To incorporate threshold effects for predicting evacuation probability, we propose the modified logistic regression models that integrate threshold effects. In particular, for each predictor $x_i$, after its threshold, $a_i$ is detected (detailed description in Section 3.3), $x_i$ is regarded to have a univariate threshold effect on the response variable and the coefficients of $x_i$ in the logistic regression model vary depending on whether $x_i$ is above or below threshold $a_i$. The logistic function of the model with the univariate effect of $x_i$ is expressed as
\begin{equation}
P(y_k=1|x_k,\beta^{(i)})=\frac{1}{1+e^{-\beta_0+\sum_{j=1}^{n}\beta_jx_{k,j}+\phi_ix_{k,i}I(x_{k,i}>a_i)}},
\end{equation}
where $\beta^{(i)}=(\beta, \phi_i)^{T}$ is the set of parameters to be estimated.

In the case of two predictors $x_i$ and $x_j$, after their bivariate thresholds $b_{i}$ and $b_{j}$ are detected (also described in Section 3.3), their interaction is treated to affect the evacuation decision when they are in one of the corresponding ranges, i.e, $x_i \leq b_{i}$ and $x_j \leq b_{j}$, $x_i \leq b_i$ and $x_j > b_j$, $x_i > b_i$ and $x_j \leq b_j$, $x_i > b_i$ and $x_j > b_j$. The logistic function of the model with bivariate effect of $x_i$ and $x_j$ is written as
\begin{equation}
P(y_k=1|x_k,\beta^{(ij)})=\frac{1}{1+e^{-\beta_0+\sum_{m=1}^{n}\beta_mx_{k,m}+\phi_{ij}\gamma_{k,i}\cdot \gamma_{k,j}}},
\end{equation}
where $\gamma_i \cdot \gamma_j$ is the bivariate threshold effect that fulfills
\begin{eqnarray*}
    \alpha_{i} \cdot \alpha_{j}=\left\{
    \begin{array}{rcl}
    x_{i} \cdot x_{j} & &{\gamma_{i} \leq b_{i}, \gamma_{j} \leq b_{j}}\ or\  \gamma_{i} > b_{i}, \gamma_{j} \leq b_j\ or\ \gamma_{i}>b_i, \gamma_{j} \leq b_j\ or\ \gamma_{i}>b_{i}, \gamma_{j}>b_{j}  \\
    0 & &\rm{otherwise}\\
    \end{array}\right.
\end{eqnarray*}
and $\beta^{(ij)}=(\beta, \phi_{ij})^{T}$ is the set of parameters to be estimated.

It is worth noting that not every threshold effect has statistical significance. Only those that are statistically significant are meaningful and should be included in the final model to prevent reaching incorrect conclusions based on random fluctuations in the data. The criteria and process for selecting significant effects are thoroughly outlined in section 3.4. The logistic function of the enhanced model (ELR) incorporating all significant threshold effects is represented as:
\begin{equation}
P(y_k=1|x_k,\beta_{ELR})=\frac{1}{1+e^{-\beta_0+\sum_{i=1}^{n}\beta_ix_{k,i}+\sum_{j=1}^{l}\phi_{uj}x_{k,uj}I(x_{k,uj}>a_{uj})+\sum_{m=1}^{q} \phi_{bm1bm2}\gamma_{k,bm1}\cdot \gamma_{k,bm2}}},
\end{equation}
where $\left(x_{u1},x_{u2}, \cdots, x_{ul}\right)^{T}$ is the set of predictors that have significant univariate threshold effects, $l$ is the number of significant univariate threshold effects, $\left(\left(\gamma_{b11},\gamma_{b12}\right), \cdots, \left(\gamma_{bq1},\gamma_{bq2}\right) \right)^{T}$ is the set of pairs of predictors that have significant bivariate threshold effects, $q$ is the number of significant bivariate threshold effects, $\beta_{ELR}=(\beta,\phi_{u1},\phi_{u2}, \cdots, \phi_{ul},$\\$\phi_{b11b12}, \cdots, \phi_{bq1bq2})^{T}$ is the set of parameters to be estimated.

\subsection{Detecting Threshold Effects With Low-depth Decision Trees}
Decision Tree is a tree-structured model utilized for regression and classification, in which the Classification and Regression Tree (CART) algorithm is more popular. In each iteration of the CART algorithm, the decision node and its corresponding splitting value minimize the Gini Index (the impurity level of the observations). The dataset is split into two subsets according to the splitting value, making the cases within respective resulting partitions more homogeneous. This splitting procedure is repeatedly carried out in CART until a pre-specified condition is met (e.g., the maximal depth of the tree is reached) or the Gini Index can no longer decrease. The CART algorithm detects optimal threshold values and facilitates capturing interactions through binary splits, making it suitable for automatically uncovering nonlinear relationships. Moreover, compared with non-pruned decision trees (multi-layer decision trees), low-depth decision trees have fewer splits and are more robust. Hence, we adopt a similar approach as \citet{DUMITRESCU20221178} by leveraging the low-depth CART in our study as it is a well-suited method for threshold effects detection.

\begin{figure}[H]
     \centering
     \begin{subfigure}{1.0\textwidth}
         \centering
         \includegraphics[width=0.55\textwidth]{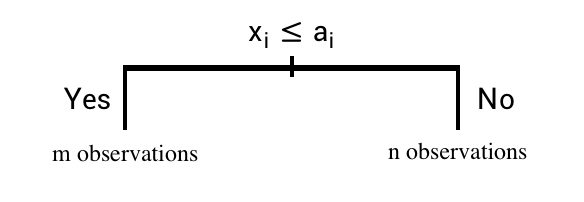}
         \caption{Univariate Threshold Effect Detection}
         \label{fig:univariate}
     \end{subfigure}\\
     \begin{subfigure}{1.0\textwidth}
         \centering
         \includegraphics[width=0.55\textwidth]{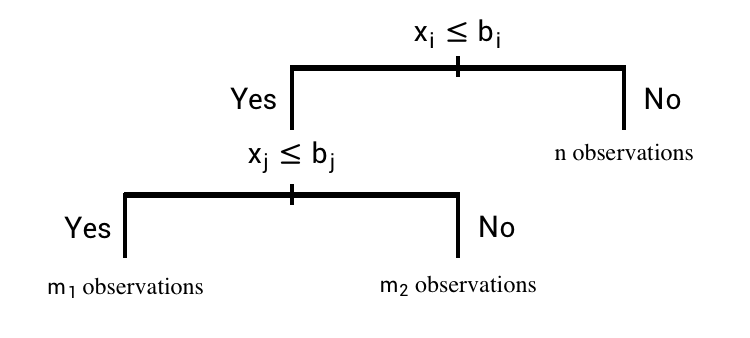}
         \caption{Bivariate Threshold Effect Detection}
         \label{fig:bivariate}
     \end{subfigure}
     \caption{Examples of Decision Trees for Threshold Effects Detection}
     \label{fig:four type of trees}
\end{figure}

For detecting the univariate threshold effect of $p$ variables, $p$ one-layer decision trees are built. Each tree consists of a single candidate node, which is one of the $p$ variables. The splitting value of the root node of each tree is selected as the threshold. The $i_{th}$ decision tree with the variable $\textbf x_i$ as the candidate node is taken as an example to explain the one-layer decision tree in detail. As depicted in Figure \ref{fig:univariate}, the tree comprises one root node ($\textbf x_i$) and two leaf nodes, each containing $m$ and $n$ observations, respectively. With the threshold observed to be $a_i$, the contribution of $\textbf x_i$ to the response variable is considered to be different when $\textbf x_i$ is below and above $a_i$.

Bivariate threshold effects are detected by running decision trees with only two variables as the candidate variables. Shapes of decision trees for threshold effect detection may differ depending on the way of splitting. The decision tree with the variable $\textbf{x}_{i}$ and $\textbf{x}_{j}$ is taken as an example for illustration. As shown in Figure \ref{fig:bivariate}, the decision tree is a two-layer tree with three leaf nodes, each with $m_1$, $m_2$, $n$ observations, respectively. $\textbf{x}_i$ is the decision node of the first iteration, and $\textbf{x}_j$ participates in the second splitting iteration. The left and middle leaf node observations are regarded to have bivariate threshold effects. That is, when $x_i \leq b_i$ and $x_j \leq b_j$, or $x_i \leq b_i$ and $x_j > b_j$, the interaction term, $\textbf{x}_i \cdot \textbf{x}_j$, affects the evacuation decision.

Some variables $\textbf{x}_i$ may have high feature importance and hence participate in the first two splitting iterations in the decision tree. For detecting their bivariate threshold effects with other variable $\textbf{x}_j$, three-layer decision trees are also applied where $\textbf{x}_i$ is the decision node in the first two iterations and $\textbf{x}_j$ participates in the third splitting iteration. We did not consider trees with more than three layers to avoid the presence of leaf nodes containing too few observations.

\subsection{Significant Threshold Effects Selection}
To examine whether the threshold effects detected by decision trees are significant, we conduct the likelihood ratio test (LRT). LRT is used to test the goodness-of-fit between two models, where one model nests the other. In this paper, LRT is applied to evaluate whether each logistic regression model with a univariate effect of $x_i$, or each logistic regression model with a bivariate effect of $x_i$ and $x_j$ fits our dataset significantly better than the original model. For each model that includes the univariate threshold effect, the likelihood ratio is
\begin{equation}
LR^{(i)}=-2[L(\hat{\beta})-L(\hat{\beta}^{(i)})].
\end{equation}
For each model that includes the bivariate threshold effect, the likelihood ratio is
\begin{equation}
LR^{(ij)}=-2[L(\hat{\beta})-L(\hat{\beta}^{(ij)})].
\end{equation}
The likelihood ratio approximately follows the $\chi^{2}$ distribution, and since only one additional parameter is added to each new regression model, the degree of freedom is one. When the $p$-value of the Likelihood Ratio Test is less than 0.01, the logistic regression model with either a univariate or a bivariate threshold effect is considered to be superior to the original model, and the two models are deemed statistically different. If the logistic regression model with a univariate threshold effect of $x_i$ is statistically better than the original model, and the $p$-values of both $x_i$ and $x_iI(x_i>a_i)$ are less than 0.01, then the univariate threshold effect of $x_i$ is considered to be significant. If the logistic regression model with a bivariate threshold effect of $x_i$ and $x_j$ has the $p$-value of LRT smaller than 0.01, and the $p$-value of $\gamma_i \cdot \gamma_j$ is smaller than 0.01, then the bivariate threshold effect of $x_i$ and $x_j$ is deemed significant.

\section{Data}
In our study, the survey data collected in the aftermath of Hurricanes Katrina and Rita and analyzed by a linear method \citep{huang2017multistage} is selected for the following nonlinearity detection and model construction.
\subsection{Hurricane Rita \& Katrina}
Hurricane Katrina was one of the most destructive hurricanes that caused a significant number of fatalities. On August 28, 2005, when Katrina moved to the Gulf of Mexico, it intensified to Category 5 from a Category 3 hurricane. Over the next three days, it gradually shifted from the southwest to the north \citep{2005a}. On August 27, a hurricane watch was issued at 10:00 a.m. Central Daylight Time (CDT) by the National Hurricane Center (NHC), and a hurricane warning was issued 13 hours later when the intensity of Katrina reached Category 3 and was still strengthening. On August 29, with the intensity of Category 3, Katrina made landfall near Buras and Louisiana; the storm surge was 2.4 to 6.7 meters (8 to 22 feet).

In September 2005, Hurricane Rita swiped Texas \citep{2005b}. A hurricane watch was issued by NHC at 4:00 p.m. CDT on September 21, and a hurricane warning was issued 18 hours later when Rita reached Category 5. The storm First threatened Corpus Christi, Texas, then turned eastward to Galveston, Texas. At 2:38 a.m. CDT on September 24, with the intensity of Category 3, it made landfall close to Sabine Pass, Texas, and Johnson Bayou, Louisiana; the storm surge was 1.2 to 2.1 meters (4 to 7 feet).

\subsection{Data Collection Procedures}
Two mail surveys that began four months after Hurricane Katrina were conducted by the Hazard Reduction and Recovery Center (HRRC) of Texas A\&M University. Two Louisiana parishes, Jefferson and St. Charles, were included in the Katrina survey, while the Rita survey covered seven Taxes counties---two coastal counties (Jefferson and Orange) and three inland counties (Newton, Jasper, and Hardin) from the Sabine study area (SSA), one coastal county (Galveston) and one inland county (Harris) from the Houston–Galveston study area (GSA).

When selecting survey takers, a disproportionate stratified sampling procedure was adopted. With the expected number of respondents to be 200 in each parish and county, and a 50\% expected response rate, the questionnaire was mailed to 800 households in Louisiana and 2800 in Taxes. Steps for questionnaire distribution and recovery referred to \citet{dillman2011mail} procedure: Each selected household received a packet including a cover letter, a questionnaire, and a stamped and self-addressed envelope. The family which did not return a completed questionnaire within two weeks would receive a reminder postcard. It would continue receiving replacement packets every two weeks until it finished the questionnaire or has already been sent three replacement packets and one reminder postcard.

270 of the 800 households taking the Katrina survey returned valid questionnaires, and the response rate was 39.9\% (37\% in Jefferson and 43\% in St. Charles). 1007 of the 2800 households participating in the Rita survey sent back completed questionnaires, and the response rate was 41.8\% (all the counties had similar response rates). Cases from two surveys had been pooled together after a homogeneity test \citep{huang2017multistage}.

\subsection{Treatment of Missing Data}
558 (42.7\%) of 1277 observations have missing data. Although the missing rates of most of the variables are lower than 5\%, these missing values may significantly affect the results. Hence, the missing values have been replaced by the expectation-maximum (EM) algorithm before entering into further analyses.

\subsection{Variable Description}
Our study only focused on demographic, geographic, and resource-related variables. Psychological variables were omitted from the model development (but were included in LR). The updated dataset contains 1277 observations and 15 variables, one as the response variable and 14 as predictors. The response variable is $\textbf{EvaDec}$ (Evacuation Decision). For the values of $\textbf{EvaDec}$, 0 is for ``To Stay at Home,'' the proportion is 17.07\%, 1 is for ``To Evacuate,'' and the proportion is 82.93\%. Except for the binary predictors (Female, White, Married, and HmOwn), all the predictors are treated as continuous. The predictor name, description, type, value or scale, and the meaning of each value are illustrated in Table \ref{tab:variables}.

For the values of the predictor $\textbf{RiskArea}$, 0 is for Barrier Islands, 1 is for locations prone to Category 1 or 2 hurricanes (Risk Areas 1 and 2 for SSA, Zip-Zone A for GSA, and Phase I for Louisiana State), 2 is for locations prone to Category 3 hurricanes (Risk Area 3 for SSA, Zip-Zone B for GSA, and Phase II for Louisiana State), 3 is for locations prone to Category 4 or 5 hurricanes (Risk Areas 4 and 5 for SSA, Zip-Zone C for GSA, and Phase III for Louisiana), and 4 is for locations farther inland. The larger the value, the farther the respondent is from the coast.

\begin{table}[H]
\caption{Predictors included in the study}\label{tab:variables}
\resizebox{\columnwidth}{!}{%
\begin{tabular}{llllllll}
\hline
Variable & Description & Mean & Std. & Min & Max & Proportion & Category \\\hline
Female & Gender & / & / & / & / & \begin{tabular}[c]{@{}l@{}}0: Male (48.63\%)\\ 1: Female (51.37\%)\end{tabular} & Demographic \\
White & Race & / & / & / & / & \begin{tabular}[c]{@{}l@{}}0: Others (22.47\%)\\ 1: White (77.53\%)\end{tabular} & Demographic \\
Married & Marital Status & / & / & / & / & \begin{tabular}[c]{@{}l@{}}0: Others (30.62\%)\\ 1: Married (69.38\%)\end{tabular} & Demographic \\
HmOwn & House Ownership & / & / & / & / & \begin{tabular}[c]{@{}l@{}}0: Others (12.61\%)\\ 1: Own House (87.39\%)\end{tabular} & Demographic \\
Age & Age & 53.53 & 15.08 & 19.00 & 94.00 & / & Demographic \\
HHSize & Household Size & 2.84 & 1.57 & 1.00 & 17.00 & / & Demographic \\
Edu & Education Years & 13.98 & 2.42 & 9.00 & 18.00 & / & Demographic \\
Income & Annual Income & 38088.00 & 12660.93 & 15000.00 & 53398.00 & / & Demographic \\
RiskArea & Risk Area & / & / & / & / & \begin{tabular}[c]{@{}l@{}}0: Barrier Island\\ 1: Risk Area A or 1 and 2\\ 2: Risk Area B or 3\\ 3: Risk Area C or 4 and 5\\ 4: Inland\end{tabular} & Geographic \\
RegVeh & \begin{tabular}[c]{@{}l@{}}Registered Vehicle \\ Number\end{tabular} & 2.15 & 0.93 & 0.000 & 9.00 & / & Resource-related \\
EvaVeh & \begin{tabular}[c]{@{}l@{}}Estimated number of\\ Vehicle to Take in the \\ Evacuation\end{tabular} & 1.412 & 0.70 & 0.00 & 5.00 & / & Resource-related \\
EvaTrail & \begin{tabular}[c]{@{}l@{}}Estimated number of \\Trailers to Take in the \\ Evacuation\end{tabular} & 0.12 & 0.32 & 0.00 & 2.00 & / & Resource-related \\
EvaCost & \begin{tabular}[c]{@{}l@{}}Estimated Cost for \\ Evacuation\end{tabular} & 1178.00 & 1875.145 & 0.00 & 41150.00 & / & Resource-related\\\hline
\end{tabular}%
}
\end{table}

\section{Results}
\subsection{Model Performance Comparison}
To fairly evaluate the model performance, we split the dataset into a 9:1 ratio for training and testing, with the training set used for threshold detection and model fitting and the test set used to assess model performance and determine if the enhanced logistic regression (ELR) model with all significant threshold effects outperforms the two baseline models: One ignoring nonlinear effects and the other incorporating psychological variables.  The four measures we chose for model performance measurement and comparison are: $R^{2}$ and Adjusted $R^{2}$ for model fitting ability; Accuracy, Precision, Recall, F1 score, and AUC for model prediction ability.

$R^{2}$ and Adjusted $R^{2}$ measure the proportion of variance in the response variable explained by the predictors. Both of them take values from 0 to 1. Higher $R^{2}$ and Adjusted $R^{2}$ show that the regression model explains the observed data well. For the response variable $y_i$, $i$=$1,2, \cdots, n$, $n$ is the number of observations, $R^{2}$ is defined as: 
\begin{equation}
R^2=\frac{SS_{Regression}}{SS_{Total}}=\frac{\sum_{i=1}^{n}(\hat{y_i}-\bar{y})^2}{\sum_{i=1}^{n}(y_i-\bar{y})^2},
\end{equation}
where $SS_{Regression}$ is the sum of squares due to the regression and $SS_{Total}$ is the total sum of squares, $\hat{y_{i}}$ is the predicted value of $y_{i}$ and $\overline{y}$ is the mean of the response values for all observations. Adjusted R$^{2}$ is an adjusted version of R$^{2}$, which considers the degree of freedom. It is defined as:
\begin{equation}
Adjusted\ R^2=1-\frac{(1-R^2)(n-1)}{n-p-1},
\end{equation}
where $p$ is the number of predictors.

To evaluate the prediction results of our model with all significant threshold effects, we also compared the models' accuracy, precision, recall, F1 score, and AUC. Accuracy measures the proportion of correctly predicted observations over the total observations. It is defined as:
\begin{equation}
Accuracy=\frac{TP+TN}{TP+TN+FP+FN},
\end{equation}
where TP (True Positive) is the number of ``to evacuate'' correctly predicted as ``to evacuate,'' TN (True Negative) is the number of ``to stay at home'' correctly predicted as ``to stay at home,'' FP (False Positive) is the number of ``to stay at home'' wrongly predicted as ``to evacuate,'' and FN (False Negative) is the number of ``to evacuate'' wrongly predicted as ``to stay at home.'' Precision measures the proportion of correctly identified positive responses out of all the positive predictions made. It is defined as:
\begin{equation}
Precision=\frac{TP}{TP+FP}.
\end{equation}
Recall measures the proportion of correctly identified positive responses out of all actual positive responses. It is defined as:
\begin{equation}
Recall=\frac{TP}{TP+FN}.
\end{equation}
F1 score is the harmonic mean of precision and recall. It is defined as:
\begin{equation}
F1=\frac{2\cdot \frac{TP}{TP+FN}\cdot \frac{TP}{TP+FP}}{\frac{TP}{TP+FN}+\frac{TP}{TP+FP}}.
\end{equation}
\noindent AUC is the area under the ROC (Receiver Operating Characteristics) curve plotted with True Positive Rate (TPR) against False Positive Rate (FPR). It also takes values from 0 to 1. When AUC is close to 1, the model is better in distinguishing between respondents with ``to evacuate'' and ``to stay at home.''

Table \ref{tab:performance} displays the statistical measures of the performance of four models: the baseline logistic regression (LR) model with the demographic, geographic, and resource-related variables as predictors, the baseline logistic regression (LR) model with psychological variables built by \citet{huang2017multistage}, the enhanced original logistic regression (ELR) model with significant univariate threshold effects, and the enhanced original logistic regression (ELR) model with all significant univariate and bivariate threshold effects.

\begin{table}[H]
\caption{Model Performance Comparison}\label{tab:performance}
\resizebox{\columnwidth}{!}{%
\begin{tabular}{llllllll}
\hline
 & \multicolumn{2}{l}{\centering In-Sample Performance} & \multicolumn{5}{l}{\centering Out-of-Sample Performance} \\\hline
Model & $R^2$ & Adj $R^2$ & Accuracy & Precision & Recall & F1 Score & AUC\\\hline
Baseline LR & 0.1141 & 0.1032 & 0.7734 & 0.7734 & 0.9904 & 0.8722 & 0.5000 \\
\begin{tabular}[c]{@{}l@{}} Baseline LR w/ Psychological Variables\end{tabular} & 0.3025 & 0.2882 & 0.8516 & 0.8571 & 0.9697 & 0.9100 & 0.7090 \\
\begin{tabular}[c]{@{}l@{}}ELR w/ Significant Univariate Threshold Effects\end{tabular} & 0.5363 & 0.5293 & 0.8750 & 0.8878 & 0.9596 & 0.9223 & 0.7729 \\
\begin{tabular}[c]{@{}l@{}}\textbf{ELR w/ All Significant Threshold Effects}\end{tabular} & \textbf{0.8316} & \textbf{0.8285} & \textbf{0.9375} & 
\textbf{0.9333} &
\textbf{0.9899} &
\textbf{0.9608} & \textbf{0.8743}\\\hline
\end{tabular}%
}
\end{table}

As shown in Table \ref{tab:performance}, the model without psychological variables performs the worst since the linear contributions of a majority of predictors are not significant. Besides, some significant effects of psychological variables are not included. For example, \citet{huang2017multistage} found that people's expectations of wind impacted and evacuation impediments, and the extent to which they took official warnings into account significantly contributed to their evacuation decisions. However, the enhanced logistic regression model (ELR) with the significant univariate threshold effects already has outperformed the model with psychological variables from both model fit and prediction ability: $R^{2}$ and Adjusted $R^{2}$ increase from 0.3 to 0.5; the growth rate of Accuracy, precision and F1 score is about 3.0\%, 3.5\%, and 1.5\%, respectively; the growth rate of AUC is about 18\%. It can be explained as people's risk perception being influenced by their demographic and resource-related features as well as their locations to some extent. Once we correctly detect the nonlinear effects of those variables, the model's predictive power can be as competitive as the model with risk perceptions (psychological variables). The model with all significant threshold effects outperforms all other models: $R^{2}$ and Adjusted $R^{2}$ increase to 0.8. Accuracy, precision, F1 score, and AUC are larger than those of ELR with significant univariate threshold effects. It indicates that the model performance is further improved when the interaction terms are added to the model.

\subsection{Significant Threshold Effects and Interpretations}
Our study focuses on the univariate threshold effects of all continuous predictors, as well as the bivariate threshold effects of variables from different categories, i.e., bivariate threshold effects of one demographic/geographic variable and one resource-related variable, as we have discussed in the Method section. The model comparison results have shown that the enhanced logistic regression model (ELR) outperforms the baseline models and can be utilized for evacuation decision prediction and analysis. We applied the ELR model on the complete dataset to better understand evacuation behavior during Hurricanes Katrina and Rita. In Table \ref{tab:thresholds}, the first four are significant univariate threshold effects, whereas the others are significant bivariate threshold effects. Table \ref{tab:regression} is the result of the logistic regression including all significant thresholds. Based on the results in Table \ref{tab:thresholds} and Table \ref{tab:regression}, we had the following observations and interpretations.

\begin{table}[H]
	\caption{Significant Thresholds and $p$-values of LRT}\label{tab:thresholds}
	\begin{center}
		\begin{tabular}{l l l l l}
		\hline
			Predictor1 & Threshold1 & Predictor2 & Threshold2 & $p$-value (LRT) \\\hline
			HHSize & $>2.39$ & / & / & $<0.01$\\
			RegVeh & $>2.01$ & / & / & $<0.01$\\
			EvaVeh & $>1.00$ & / & / & $<0.01$\\
			EvaCost & $>704.03$ & / & / & $<0.01$\\
			RiskArea & $>3.45$ & EvaCost & $>704.03$ & $<0.01$\\
			HHSize & $\leq15.00$ & RegVeh & $>2.99$ & $<0.01$\\
			HHSize & $\leq13.50$ & EvaVeh & $>2.00$ & $<0.01$ \\
			Edu & $>10.33$ & EvaCost & $\leq 511.22$ & $<0.01$
			\\\hline
		\end{tabular}
	\end{center}
\end{table}

Gender can be regarded as having a generalized linear effect on the evacuation decision. Compared with men, women are more likely to evacuate ($\beta$ = 1.535). Marital status is considered to affect the decision-making in a generalized linear manner. Married people are more likely to evacuate ($\beta$ = 1.480).

For the household size, when it is smaller than 2.39, it negatively contributes to the evacuation decision ($\beta$ = -0.984). However, when it is above 2.39, its negative impact decreases ($\beta^{'}$ = -0.746). Although households with more than two people are more likely to contain elders, who have more evacuation impediments (e.g., disability and massive medical equipment), it is also more likely to include children. \citet{gladwin1997warning} raised that families with children are more likely to evacuate. Both impacts cancel each other out to some extent. It might be why the effect of household size decreases when it is above 2.39. For the number of registered vehicles, when it is smaller than 2.01, it has a positive contribution to the decision-making ($\beta$= 0.681). When it is above 2.01, its influence becomes more negligible ($\beta^{'}$= 0.055). It can be explained that, although more registered vehicles imply more available resources for evacuation, it follows the law of diminishing marginal utility that households would only drive a necessary number of cars in their evacuation trip. The estimated number of vehicles to take in the evacuation positively affect the evacuation decision when smaller than 1.00 ($\beta$ = 3.121). When it is above 1.00, it negatively contributes to the decision-making ($\beta^{'}$ = -1.694). It indicates that people who were able to take a car for evacuation are more likely to evacuate, whereas additional vehicle requirements could be a burden on households' evacuation. When the estimated evacuation cost is smaller than 704.03, it positively contributes to the evacuation decision ($\beta$ = 0.003). When it is above 704.03, its influence becomes negligible ($\beta^{'}$ = 0.000). It implies households had a plan on the expense scale of their evacuation trip. Comprehensively speaking, the results of evacuation vehicle requirements and expected evacuation costs indicate that the increased resource requirements and costs would hesitate households' ultimate evacuation decisions.

\begin{table}[ht]
	\caption{Logistic regression with all significant threshold effects}\label{tab:regression}
	\begin{center}
            \footnotesize
	    \begin{threeparttable}
		\begin{tabular}{l l l l l}
		\hline
			Variable & Estimate & Std. Error & $z$ value & $p$-value \\\hline
			Intercept & 1.383 & 1.642 & 0.842 & 0.400\\
			Age & -0.009 & 0.012 & -0.707 & 0.480\\
			\textbf{Female} & \textbf{1.535} & \textbf{0.356} & \textbf{4.306} & \textbf{$<$0.001$^{***}$}\\
			White & -0.190 & 0.367 & -0.517 & 0.605\\
			\textbf{Married} & \textbf{1.480} & \textbf{0.428} & \textbf{3.454} & \textbf{$<$0.001$^{***}$}\\
			\textbf{HHSize} & \textbf{-0.984} & \textbf{0.366} & \textbf{-2.685} & \textbf{0.007$^{**}$} \\
			Edu & 0.045 & 0.073 & 0.614 & 0.539\\
			Income & 0.000 & 0.000 & -1.053 & 0.292\\
			HmOwn & -0.312 & 0.198 & -0.638 & 0.523\\
			RiskArea & -0.329 & 0.515 & -1.576 & 0.115\\
			\textbf{RegVeh} & \textbf{0.681} & \textbf{0.360} & \textbf{1.919} & \textbf{0.055\textasciitilde}\\
			\textbf{EvaVeh} & \textbf{3.121} & \textbf{0.547} & \textbf{5.710} & \textbf{$<$0.001$^{***}$}\\
			EvaTrail & 0.229 & 0.607 & 0.377 & 0.706\\
			\textbf{EvaCost} & \textbf{0.003} & \textbf{0.001} & \textbf{3.036} & \textbf{0.002$^{**}$}\\
			HHSize($>2.39$) & 0.238 & 0.226 & 1.055 & 0.291\\
			\textbf{RegVeh($>$2.01)} & \textbf{-0.626} & \textbf{0.231} & \textbf{-2.713} & \textbf{0.007$^{**}$}\\
			\textbf{EvaVeh($>$1.00)} & \textbf{-4.815} & \textbf{0.468} & \textbf{-10.288} & \textbf{$<$0.001$^{***}$}\\
			\textbf{EvaCost($>$704.03)} & \textbf{-0.003} & \textbf{0.001} & \textbf{-2.756} & \textbf{0.006$^{**}$}\\	RiskArea($>3.45$)$\cdot$EvaCost($>704.03$) & 0.000 & 0.000 & -1.409 & 0.159\\	\textbf{HHSize($\leq$15.00)$\cdot$RegVeh($>$2.99)} & \textbf{0.157} & \textbf{0.077} & \textbf{2.031} & \textbf{0.042$^{*}$}\\
			\textbf{HHSize($\leq$13.50)$\cdot$EvaVeh($>$2.00)} & \textbf{0.969} & \textbf{0.120} & \textbf{8.047} & \textbf{$<$0.001$^{***}$}\\
			\textbf{Edu($>$ 10.33)$\cdot$EvaCost($\leq$511.22)} & \textbf{0.000} & \textbf{0.000} & \textbf{3.131} & \textbf{0.002$^{**}$}
			\\\hline
		\end{tabular}
		\begin{tablenotes}
		\item[1] $p$-value: $<$0.001 *** $<$0.01 ** $<$0.05 * $<$0.10 \textasciitilde
		\end{tablenotes}
		\end{threeparttable}
	\end{center}
\end{table}

The interaction term of household size ($\leq$15.00) and the number of registered vehicles ($>$2.99) positively affects the evacuation decision. It means that when the household size increases, the coefficient of registered vehicles increases. For a larger family, the number of registered vehicles has a more positive contribution to decision-making. It is rational because a larger family usually needs more evacuation tools. As a result, fewer vehicles are left at home, which means that the larger family has more minor evacuation impediments. The interaction term of household size ($\leq$13.5) and the number of vehicles to take in the evacuation ($>$2.00) positively influence the evacuation decision. It indicates that when the household size increases, the coefficient of vehicles to take in the evacuation increases. For a large family, the number of cars to take in the evacuation positively affects decision-making. Besides, the interaction term of education ($>$10.33) and evacuation cost ($\leq$511.22) has a positive contribution to the decision-making. It shows that when the education year increases, the coefficient of estimated evacuation cost also increases. 

\section{Discussion}

As shown in Table 4, the proposed ELR model significantly outperforms the traditional approaches (i.e., baseline linear regression models with or without psychological variables) in terms of both model fit and prediction accuracy. This is due to ELR's capability to automatically detect nonlinearities (univariate threshold effects) and interactions (bivariate threshold effects), assess their statistical significance, and incorporate significant threshold effects into the logistic regression models. Furthermore, the outstanding performance of ELR suggests that once the complex mechanism in which social and household contextual variables affect evacuation decisions can be captured by the model, accurate predictions can be achieved based on these objective and easily-accessible variables, thereby reducing the reliance on psychological variables. Therefore, we can conclude that ELR has improved the timeliness and accuracy of the evacuation decision model without collecting householders' risk perceptions and other psychological variables.

 For high-stakes decision-making, such as evacuation planning, more accurate predictions are required to give emergency managers reliable evidence into evacuation zone division, potential risks, and other relevant goals \citep{dash2007evacuation}. Additionally, the models or systems should be transparent to be trusted and adopted by experts \citep{kim2015interactive}. As a result, in such applications, accurate interpretable machine learning models are probably preferred over the black-box counterparts, making ELR a promising modeling technique. For example, ELR has high prediction capability as well as preserving the transparency and intrinsic interpretability of the logistic regression model.
 Thus, ELR may be directly used by emergency managers to forecast evacuation traffic demand and identify households requiring assistance in evacuations. 

Simulation is also essential for high-stakes decision-making domains, particularly in providing precautions for disaster-prone areas which are either remote or have not yet experienced a disaster \citep{ramaguru2019human}. In order to generate and analyze simulations effectively, human behavior must be considered by supporting technologies \citep{lee2004study, kim2004establishing, zhao2021using}, among which AI has shown great potential for helping existing traffic simulators gain more realistic outputs \citep{zhao2021using} and rapidly extracting useful knowledge \citep{sun2020applications}. Therefore, it is of high potential to leverage ELR to assist the development of the next-generation AI-based evacuation simulation tools, due to its high prediction accuracy, straightforward model formulation (Eqn. (5)), minimal need for model specification, and reliance only on readily simulated social and household contextual data. 

Additionally, this study also brings to light some new findings related to the underlying relationships between social and household contextual variables and evacuation decisions. Specifically, we uncovered the nonlinear patterns of demographic variables and the diminishing marginal effects of resource-related variables. The effect of \textit{HHSize} (Household Size) is statistically insignificant when treated as linear, but attains significance after its univariate threshold effect is detected. When the value of each resource-related variable (i.e., \textit{RegVeh}, \textit{EvaVeh}, \textit{EvaCost}) exceeds the specific threshold, its marginal effect decreases. This implies that when available resources reach certain limits, their impacts on people's evacuation decisions become minimal.

\section{Conclusion}

In this study, we developed a new methodology to predict hurricane evacuation decisions with an interpretable machine learning approach, i.e., an enhanced logistic regression model (ELR) that is assisted by low-depth decision trees to detect nonlinear effects. The empirical results have verified the conjecture that the ELR can incorporate nonlinearities determined by a sequence of univariate and bivariate thresholds among demographic and resource-related variables that could well predict households’ evacuation decisions as psychological models. This study thus contributes to the protective action behavioral theories by showing that the predictors of household and social contexts including resource availabilities, social vulnerabilities, and social capital have significant effects (including linear and nonlinear effects) in predicting householders' evacuation decisions.

Admittedly, this study has some limitations and future work can be done to further explore the application of interpretable machine learning models in predicting hurricane evacuation decisions. For example, in our study, we used one dataset to compare the performance of multiple models and test the validity of the proposed ELR. In future studies, more datasets from different evacuation events could be used to examine whether ELR can be generalized to different hurricane events. Additionally, only univariate and between-category bivariate threshold effects are considered in this proposed model; in future work, within-category bivariate threshold effects can also be considered and non-pruned decision trees (e.g., CART) may also be taken into account to check whether adding highly nonlinear multivariate threshold effects can further improve the model’s prediction capability or just increase the model's overfitting problem.

\section*{Author contributions}
The authors confirm contributions to the paper as follows: study conception and design: Sun and Zhao; data curation and model implementation: Sun and Huang; analysis and interpretation of results: Sun, Huang, and Zhao; draft manuscript preparation: Sun, Huang, and Zhao. All authors reviewed the results and approved the final version of the manuscript.

\section*{Acknowledgements}
This material is based upon work supported by the National Science Foundation under Grant No. 2303578, 2303579, 0527699, 0838654, 1212790 and by an Early-Career Research Fellowship from the Gulf Research Program of the National Academies of Sciences, Engineering, and Medicine. Any opinions, findings, and conclusions or recommendations expressed in this material are those of the authors and do not necessarily reflect the views of the National Science Foundation or the Gulf Research Program of the National Academies of Sciences, Engineering, and Medicine. The authors would like to thank Drs. Michael Lindell and Carla Prater for sharing the data with the research team. 

\bibliographystyle{apalike}
\bibliography{sample}

\end{document}